\def\year{2020}\relax
\documentclass[letterpaper]{article} 
\usepackage{aaai20}  
\usepackage{times}  
\usepackage{helvet} 
\usepackage{courier}  
\usepackage[hyphens]{url}  
\usepackage{graphicx} 
\usepackage{graphicx}  
\usepackage{amsmath}
\usepackage{amsthm}
\usepackage{mathtools}
\usepackage{dsfont}
\usepackage{xcolor}
\usepackage{amsfonts}

\usepackage{bm}
\newtheorem{theorem}{Theorem}

\newtheorem{example}{Example}
\newtheorem{remark}{Remark}
\newtheorem{defn}{Definition}

\usepackage[utf8]{inputenc}
\usepackage[english]{babel}
\title{Attribute noise robust binary classification}
\author{Aditya Petety$^1$, Sandhya Tripathi$^2$, N Hemachandra$^2$  \\ $^1$National Institute of Science Education and Research, Bhubaneshwar \\
$^1$aditya.petety@niser.ac.in \\
$^2$Indian Institute of Technology Bombay \\
$^2$\{sandhya.tripathi, nh\}@iitb.ac.in}
 
\begin{document}

\maketitle

\begin{abstract}
We consider the problem of learning linear classifiers when both features and labels are binary. In addition, the features are noisy, i.e., they could be flipped with an unknown probability. In Sy-De attribute noise model, where all features could be noisy together with same probability, we show that $0$-$1$ loss ($l_{0-1}$) need not be robust but a popular surrogate, squared loss ($l_{sq}$) is. In Asy-In attribute noise model, we prove that $l_{0-1}$ is robust for any distribution over 2 dimensional feature space. However, due to computational intractability of $l_{0-1}$, we resort to $l_{sq}$ and observe that it need not be Asy-In noise robust.
Our empirical results support Sy-De robustness of squared loss for low to moderate noise rates.
\end{abstract}
\section{Introduction}
	Quality of data is being compromised as its quantity is  getting larger. In classification setup, bad quality data could be due to noise in the labels or noise in the features. Label noise research has gained a lot of attention in last decade \cite{SastryManwani2016Book}. In contrast, feature or attribute noise is still unexplored. 
	As opposed to continuous valued attributes, noise in categorical features, particularly binary, can drastically change the relative location of a data point and significantly impact the classifier's performance. 
	
	\cite{quinlan1986effect} studied the effect of noise when the algorithms are decision trees. \cite{zhu2004classStudy,khoshgoftaar2009empirical} study attribute noise from the perspective of detecting noisy data points and correcting them. 
	
	Our major contributions lie in identifying  loss functions that are robust (or not) to attribute (binary valued) noise in Empirical Risk Minimization (ERM) framework.  This has an advantage that there is no need of either knowing the true value or cross-validating over or estimating the noise rates.
\section{Problem description} 
	Let $D$ be the joint distribution over $\mathbf{X} \times Y$, where $\mathbf{X} \in \mathcal{X} \subseteq \{-1,1\}^{n}$ and $Y \in \mathcal{Y} = \{-1,1\}$. 
	 Let the decision function be $f:\mathbf{X}\mapsto \mathbb{R}$, hypothesis class of all measurable functions be $\mathcal{H}$ and class of linear hypothesis be $\mathcal{H}_{lin}= \{ (\boldsymbol{\beta},c), \boldsymbol{\beta}\in \mathbb{R}^n, c \in \mathbb{R}: \Vert \boldsymbol{\beta} \Vert_2 \leq B\}.$ We restrict our set of hypothesis to be in $\mathcal{H}_{lin}$.
	 Let $\tilde{D}$ denote the distribution on $\tilde{\mathbf{X}}\times {Y}$ obtained by inducing noise to $D$ with $\tilde{\mathbf{X}} \in \mathcal{X} \subseteq \{-1,1\}^{n}$. The corrupted sample is $\tilde{S} := \{(\tilde{\mathbf{x}}_1,{y}_1),\ldots,(\tilde{\mathbf{x}}_m,{y}_m)\} \sim \tilde{D}^m.$ 
	 The probability that the value of $i^{th}$ attribute is flipped is given by $p_{i} = (\tilde{\mathbf{X}}=-x|\mathbf{X}=x, Y=y), i \in [n]$. 
We assume that the class/label does not change with noise in the attributes.
	
	Based on the flipping probability and the dependence between events of flipping for different attributes, we \textit{identify} two attribute noise models. If all the attribute values are flipped together with same probability $p$, then it is referred to as the \textbf{symmetric dependent attribute noise model (Sy-De)}. If each attribute $j$ flips with probability $p_j$ independently of any other attribute $k \in [n]\backslash \{j\}$, then it is referred to as the \textbf{asymmetric independent attribute noise model (Asy-In)}. Even though Sy-De attribute noise model is simple, it cannot be obtained by taking $p_i=p_j,~\forall i,j \in [n]$ in Asy-In attribute noise model. Real world example of Sy-De (or Asy-In) noisy attributes: Consider a room with many sensors connected in series (or with individual battery) measuring temperature, humidity, etc., as binary value, i.e., high or low. A power failure (or battery failures) will lead to all (or individual) sensors/attributes providing noisy observations with same (or different) probability. 
	
	We consider ERM framework for classification. A natural choice for loss function is $0$-$1$ loss, i.e., $l_{0-1}(f(\mathbf{x},y)) = \mathds{1}_{[f(\mathbf{x})y <0]}$. Bayes classifier $f^*= \arg \min_{f \in \mathcal{H}}R_{D}(f)$  and Bayes risk is $R_{D}(f^*)= \min_{f \in \mathcal{H}}R_{D}(f)$ where $R_{D}(f) = E_{D}[\mathds{1}_{[f(\mathbf{x})y <0]}]$. Corresponding quantities for noisy distribution $\tilde{D}$ are $R_{\tilde{D}}(\tilde{f}^*) = \min_{\tilde{f}\in \mathcal{H}}E_{\tilde{D}}[\mathds{1}_{[\tilde{f}(\tilde{\mathbf{x}})y <0]}]$ and $\tilde{f}^* = \arg\min_{\tilde{f}\in \mathcal{H}}R_{\tilde{D}}(\tilde{f})$.
	
	Non-convex nature of $0$-$1$ loss makes it difficult to optimize and hence convex upper bounds (surrogate losses) are used in practice. In this work, we consider the squared loss $l_{sq}(f(\mathbf{x},y)) = (y- f(\mathbf{x}))^2$,  a differentiable and convex surrogate loss function. Our restriction of hypothesis to linear class $\mathcal{H}_{lin}$ can be interpreted as a form of regularization. Expected squared clean and corrupted risks are $R_{D, l_{sq}}(f) = E_{D}[(y- f(\mathbf{x}))^2]$ and $R_{\tilde{D}, l_{sq}}(\tilde{f}) = E_{\tilde{D}}[(y- \tilde{f}(\tilde{\mathbf{x}}))^2]$. Hypothesis in $\mathcal{H}_{lin}$ minimizing these clean and corrupted risks are denoted by $f^*_{sq,lin}$ and $\tilde{f}^*_{sq,lin}$.
	Next, we define attribute noise robustness of risk minimization scheme $A_{l}$.
	\begin{defn} \label{def: robustness_def}
		Let $f^{*}_{A_l}$ and $\tilde{f}^{*}_{A_l}$ be obtained from clean and corrupted distribution $D$ and $\tilde{D}$ using any arbitrary scheme $A_l$. Then, scheme $A_l$ is said to be attribute noise robust if 
	 $$R_{D}(f^{*}_{A_l}) = R_{D}(\tilde{f}^{*}_{A_l}).$$
		 Also, $l$ is said to be an attribute noise robust loss function.
	\end{defn}
	
	\section{Attribute noise robust loss functions}
	We, first, consider Sy-De attribute noise model and present a counter example (Example \ref{exa: 0-1_1d_sy-de}) to show that $0$-$1$ loss need not be robust to Sy-De attribute noise. To circumvent this problem, we provide a positive result by showing that squared loss is Sy-De attribute noise robust with origin passing linear classifiers (Theorem \ref{thm: nDi_sq}). Our hypothesis set belongs to $\mathcal{H}_{lin}$ which could be further categorized into origin passing and non-origin passing ($c=0$ or not). Details of examples and proofs are available in Supplementary Material (SM).
	\begin{example} \label{exa: 0-1_1d_sy-de}
		Consider a population of two data points (in 1-D) $(x,y)$ as $(-1,1)$ and $(1,-1)$ with probability $0.25$ and $0.75$ with a classifier $f_{lin}(x) = bx+c$. Then, the $l_{0-1}$ optimal clean  classifier is $f_{lin}^* = (b^*,c^*)= (-1,-0.1)$ with $R_{D}(f_{lin}^*) = 0$. Also, the  $l_{0-1}$ optimal Sy-De attribute noise ($p=0.4$) corrupted classifier is $\tilde{f}^*_{lin} = (\tilde{b}^*,\tilde{c}^*) = (1,-2)$ with $R_{D}(\tilde{f}_{lin}^*) = 0.25$. Since, $R_{D}(\tilde{f}_{lin}^*) \neq R_{D}(f_{lin}^*)$, $0$-$1$ loss function need not be Sy-De attribute noise robust.
	\end{example}
	\begin{theorem} \label{thm: nDi_sq}
	Consider a clean distribution $D$ on $\mathbf{X}\times Y$ and Sy-De attribute noise corrupted distribution $\tilde{D}$ on $\tilde{\mathbf{X}}\times Y$ with noise rate $p<0.5$. Then, squared loss $l_{sq}$ with origin passing linear classifiers is Sy-De attribute noise robust,i.e.,
	\begin{small}
		\begin{equation}
R_{D}(\tilde{f}_{lin,l_{sq}}^*) = 
R_{D}({f}_{lin,l_{sq}}^*)
		\end{equation}
	\end{small}
	where $f^*_{lin,l_{sq}} = (\beta_1^*,\ldots, \beta_{n}^*)$ and $\tilde{f}^*_{lin,l_{sq}} =(\tilde{\beta}_1^*, \ldots, \tilde{\beta}_n*)$ correspond to optimal linear classifiers learnt using squared loss on clean ($D$) and corrupted ($\tilde{D})$ distribution.	
	\end{theorem}
\begin{remark}
Sy-De robustness of squared loss is an interesting result because given an attribute noise corrupted dataset, obtaining a linear classifier entails solving only a linear system of equations. (Demonstrated on UCI datasets.)
\end{remark}
Now, we consider Asy-In attribute noise model and show that $0$-$1$ loss is robust to this noise with non-origin passing classifiers when $n=2$ (Theorem \ref{thm: 2d_0-1}). As $l_{0-1}$ based ERM is computationally intractable, we consider $l_{sq}$ and present a counter example to show that $l_{sq}$ need not be Asy-In noise robust (Example \ref{exa: sq_2d_asy-in}).
	\begin{theorem} \label{thm: 2d_0-1}
		Consider a clean distribution $D$ with probabilities $\{d_1,d_2,d_3,d_4\}$ on $\mathbf{X}\times Y$ with $n=2$ (population of $2^n$ data points) and Asy-In attribute noise corrupted distribution $\tilde{D}$ on $\tilde{\mathbf{X}}\times Y$ with noise rates $p_1<0.5$ and $p_2<0.5$. Then, $0$-$1$ loss with non-origin passing linear classifiers is Asy-In attribute noise robust, i.e.,
		\begin{small}
		\begin{equation}
 R_{D}(\tilde{f}_{lin,l_{0-1}}^*) = 
 R_{D}({f}_{lin,l_{0-1}}^*)
		\end{equation}
		\end{small}
		where $f^*_{lin,l_{0-1}} = (\beta_1^*, 1, c^*)$ and $\tilde{f}^*_{lin,l_{0-1}} =  (\tilde{\beta}_1^*, 1, \tilde{c}^*)$ correspond to optimal linear classifiers learnt using $0$-$1$ loss on clean ($D$) and corrupted ($\tilde{D})$ distribution respectively.
	\end{theorem}
	\begin{example} \label{exa: sq_2d_asy-in}
	Consider a population of 3 data points (in 2-D) $(x_1,x_2,y)$ as $(-1,1,1)$, $(-1,-1,-1)$, and $(1,-1,1)$ with probabilities as $(\frac{1}{4},\frac{1}{2},\frac{1}{4})$ with a classifier $f_{lin}(\mathbf{x}) = b_1x_1+b_2x_2$. Then, $l_{sq}$ optimal clean  classifier is $f_{lin,l_{sq}}^* = (b_1^*,b_2^*)=(0.5,0.5)$ with $R_{D}(f_{lin,l_{sq}}^*) = 0.5$. Also, $l_{sq}$ optimal Asy-In attribute noise ($p_1=0.1,~p_2 = 0.2$) corrupted  classifier is $\tilde{f}_{lin,l_{sq}}^* = (\tilde{b}_1^*,\tilde{b}_2^*)=(0.4,0.3)$ with $R_{D}(\tilde{f}_{lin,l_{sq}}^*) = 0.25$. Since, $R_{D}(f_{lin,l_{sq}}^*) \neq R_{D}(\tilde{f}_{lin,l_{sq}}^*)$, squared loss need not be Asy-In attribute noise robust.  
	\end{example}
\section{Experiments}
\begin{figure}[!h]
    \centering
    \includegraphics[width=0.4\textwidth]{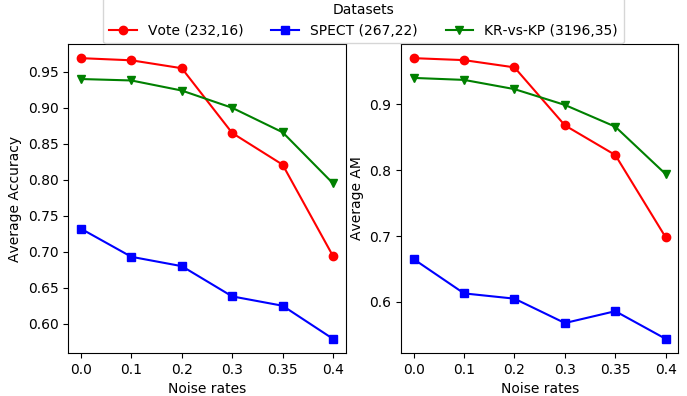}
    \caption{{ \small Test data performance of $l_{sq}$ with Sy-De attribute noise.}}
    \label{fig: real_data}
\end{figure}

Figure \ref{fig: real_data} demonstrates Sy-De attribute noise robustness of squared loss on 3 UCI datasets \cite{UCI_dataset}; details in SM. 
As SPECT dataset is imbalanced, in addition to accuracy, we also report arithmetic mean (AM). To account for randomness in noise, results are averaged over 15 trials of train-test partitioning (80-20). The low accuracy in comparison to clean classifier can be attributed to the finite samples available for learning the classifiers.

\section{Looking forward}
Our work is an initial attempt in binary valued attribute noise; an extension to general discrete valued attributes would be interesting. Asy-In attribute noise model raises some non-trivial questions w.r.t. choice of loss functions like robustness of $0$-$1$ for $n>2$, explanation for the surprising non-robustness of squared loss as compared to robustness of a difficult to deal $0$-$1$ loss, search for other surrogate loss functions that are robust. Finally, we believe that attribute dimension $n$ could have a role to play in noise robustness. 
\begin{small}
\bibliographystyle{apalike}
\bibliography{ref}
\end{small}

\newpage

\begin{center}
	\Large{\textbf{Supplementary material}}
\end{center}
\appendix
\section{Proofs}

\subsection{Proof of Theorem \ref{thm: nDi_sq}}
\begin{proof}
		Consider squared loss based clean risk given as follows:
		$$R_{D,l_{sq}}(f) = \mathbb{E}_{\mathbf{X}\times Y}[(Y- \boldsymbol{\beta}^{T}\mathbf{X})^{2}]$$ We minimize this by differentiating the expectation term and equating it to 0. $$\mathbb{E}_{\mathbf{X} \times Y}[2(Y-\boldsymbol{\beta}^{T}\mathbf{X})\mathbf{X}^{T}] = 0$$ $$\implies \mathbb{E}_{\mathbf{X} \times Y}[\mathbf{X}^{T}Y - \boldsymbol{\beta}^{T}\mathbf{X}\mathbf{X}^{T}] = 0$$ $$\implies \boldsymbol{\beta}^{*} = (\mathbb{E}[\mathbf{XX}^{T}]^{-1})^{T}\mathbb{E}[Y\mathbf{X}]$$
		Now, to obtain the optimal noisy  classifier $\tilde{f}^*_{lin,l_{sq}}$ we minimize the following squared loss based expected risk.
		\begin{eqnarray} \nonumber
		R_{\tilde{D},l_{sq}}(\tilde{f}) &=& \mathbb{E}_{\tilde{\mathbf{X}}\times Y}[(Y-\tilde{\boldsymbol{\beta}}^{T}\tilde{\mathbf{X}})^{2}] \\ \nonumber
		&=& (1-p)\mathbb{E}_{\mathbf{X}\times Y}[(Y- \tilde{\boldsymbol{\beta}}^{T}\mathbf{X})^{2}] \\ \label{eq: sq_corr_nd}
		& & + p\mathbb{E}_{\mathbf{X}\times Y}[(Y+ \tilde{\boldsymbol{\beta}}^{T}\mathbf{X})^{2}]
		\end{eqnarray}
		We minimize the corrupted risk given in equation (\ref{eq: sq_corr_nd}) by differentiating and equating to 0 as described below:
		$$(1-p)\mathbb{E}_{\mathbf{X} \times Y}[(Y - \tilde{\boldsymbol{\beta}}^{T}\mathbf{X})(-2\mathbf{X}^{T})]$$ $$ + p\mathbb{E}_{\mathbf{X} \times Y}[(Y + \tilde{\boldsymbol{\beta}}^{T}\mathbf{X})(2\mathbf{X}^{T})] = 0$$ $$\implies (p - 1)\mathbb{E}[\mathbf{X}^{T}Y] - (p - 1)\tilde{\boldsymbol{\beta}}^{T}\mathbb{E}[\mathbf{X}\mathbf{X}^{T}]$$ $$+ p\tilde{\boldsymbol{\beta}}^{T}\mathbb{E}[\mathbf{X}\mathbf{X}^{T}] + p\mathbb{E}[\mathbf{X}^{T}Y]=0$$ $$\implies (2p - 1)\mathbb{E}[\mathbf{X}^{T}Y] + \tilde{\boldsymbol{\beta}}^{T}\mathbb{E}[\mathbf{X}\mathbf{X}^{T}] = 0$$ $$\implies \tilde{\boldsymbol{\beta}}^* = (1-2p)(\mathbb{E}[\mathbf{XX}^{T}]^{-1})^{T}\mathbb{E}[Y\mathbf{X}]$$
		$$\implies \tilde{\boldsymbol{\beta}}^* = (1-2p)\boldsymbol{\beta}^*$$
		We can see that the noisy classifier $\tilde{f}^*_{lin,l_{sq}}$ is just a scaled version of the classifier, $f^*_{lin,l_{sq}}$, obtained from the clean risk. Since, for attribute noise robustness it is sufficient that $sign(\tilde{\boldsymbol{\beta}}^*\mathbf{x}) = sign(\boldsymbol{\beta}^*\mathbf{x})$, the aforementioned observation proves that the squared loss function is Sy-De attribute noise robust when $p<0.5$. 
	\end{proof}

\subsection{Proof of Theorem \ref{thm: 2d_0-1}}
\begin{proof}
We prove the robustness of $0$-$1$ loss in 2 dimension by taking an exhaustive search approach. Even though we consider a particular non-uniform distribution $\{d_1,d_2,d_3,d_4\}$ over 4 data points in the population and certain values of $p_1$ and $p_2$, the following claims hold for any arbitrary value of $d_i \in [0,1], ~~ \sum\limits_{i=1}^{4}d_i =0,~ i=1,\ldots,4$  and any pair of noise rate $p_1,p_2<0.5$. In the population, there are 16 ways in which four data points can be assigned to two classes. By symmetry, we can restrict ourselves to just these 4 cases.

The probabilities of $(\mathbf{x}^{(1)},y^{(1)})$, $(\mathbf{x}^{(2)},y^{(2)})$, $(\mathbf{x}^{(3)},y^{(3)})$, $(\mathbf{x}^{(4)},y^{(4)})$ in the distribution $D$ are taken to be $0.25, 0.33, 0.39$ and $0.03$ respectively. Here, $\mathbf{x}^{(i)} = [x_{i1},x_{i2}],~ i =1,\ldots,4$. The noise rates are $p_1 = 0.12$ and $p_2 = 0.23$. We consider a classifier of the form $f(\mathbf{x})= \beta\mathbf{x} + c$ where $\beta = [b1,1]$.

The clean $0$-$1$ risk denoted by $R$ is given as follows:
\begin{eqnarray*}
R = R_{D}(f_{lin}) = \mathbb{E}_{D}[\mathds{1}_{[(\beta \mathbf{x}+c)y <0]}]
= 0.25\mathds{1}_{[(b1 x_{11} +x_{12}+c)y_ <0]} \\ + 0.33\mathds{1}_{[(b1 x_{21} +x_{22}+c)y_2 <0]} 
+0.39\mathds{1}_{[(b1 x_{31} +x_{32}+c)y_3 <0]} \\ +0.03\mathds{1}_{[(b1 x_{41} +x_{42}+c)y_4 <0]}
\end{eqnarray*}

The noisy $0$-$1$ risk denoted by $R'$ is given as follows:
\begin{scriptsize}
\begin{eqnarray*}
R' = R_{\tilde{D}}(\tilde{f}_{lin}) = \mathbb{E}_{\tilde{D}}[\mathds{1}_{[(\tilde{\mathbf{\beta}} \tilde{\mathbf{x}}+c')y <0]}] \\
= (1-p_1)(1-p_2)\big[0.25\mathds{1}_{[(b1' x_{11} +x_{12}+c')y_ <0]}  + 0.33\mathds{1}_{[(b1' x_{21} +x_{22}+c')y_2 <0]}  \\ + 0.39\mathds{1}_{[(b1' x_{31} +x_{32}+c')y_3 <0]}  +0.03\mathds{1}_{[(b1' x_{41} +x_{42}+c')y_4 <0]} \big] \\
+p_1p_2 [ 0.25\mathds{1}_{[(-b1' x_{11} -x_{12}+c')y_ <0]}  + 0.33\mathds{1}_{[(-b1' x_{21} -x_{22}+c')y_2 <0]}  \\ + 0.39\mathds{1}_{[(-b1' x_{31} -x_{32}+c')y_3 <0]}  +0.03\mathds{1}_{[(-b1' x_{41} -x_{42}+c')y_4 <0]}  ] \\
+ (1-p_1)p_2 [0.25\mathds{1}_{[(b1' x_{11} -x_{12}+c')y_ <0]}  + 0.33\mathds{1}_{[(b1' x_{21} -x_{22}+c')y_2 <0]}  \\ + 0.39\mathds{1}_{[(b1' x_{31} -x_{32}+c')y_3 <0]}  +0.03\mathds{1}_{[(b1' x_{41} -x_{42}+c')y_4 <0]}] \\
+ p_1(1-p_2)[0.25\mathds{1}_{[(-b1' x_{11} +x_{12}+c')y_ <0]}  + 0.33\mathds{1}_{[(-b1' x_{21} +x_{22}+c')y_2 <0]}  \\ + 0.39\mathds{1}_{[(-b1' x_{31} +x_{32}+c')y_3 <0]}  +0.03\mathds{1}_{[(-b1' x_{41} +x_{42}+c')y_4 <0]}]
\end{eqnarray*}
\end{scriptsize}

To find the minimizer of the clean and corrupted risks $R$ and $R'$, we plot them (in MATLAB) as a function of $(b1,c)$ and $(b1',c')$. In the 4 cases considered below, we observed that even though the minimum value for $R$ and $R'$ are different, they have same minimizers. This implies that the clean $0$-$1$ risk corresponding to the optimal clean and noisy classifier would be same. And hence, condition for attribute noise robustness in Definition \ref{def: robustness_def} is satisfied.

Next, we provide the details for each case.
 \begin{equation*}
    \text{\textbf{Case 1}  }(\mathbf{x}^{(1)},y^{(1)}) = (\begin{bmatrix} +1 \\
    +1
    \end{bmatrix} , +1) \quad (\mathbf{x}^{(2)},y^{(2)}) = (\begin{bmatrix} +1 \\ -1
    \end{bmatrix}, -1)
\end{equation*} 
\begin{equation*}
    (\mathbf{x}^{(3)},y^{(3)}) = (\begin{bmatrix} -1 \\ -1
    \end{bmatrix}, +1) \quad (\mathbf{x}^{(4)},y^{(4)}) = (\begin{bmatrix} -1 \\ +1   \end{bmatrix}, -1)
\end{equation*} 
The set of classifiers obtained by minimizing the 0-1 risk is given in Fig \ref{fig: case1_Clean}.
\begin{figure}[!h]
    \centering
\includegraphics[scale=0.4]{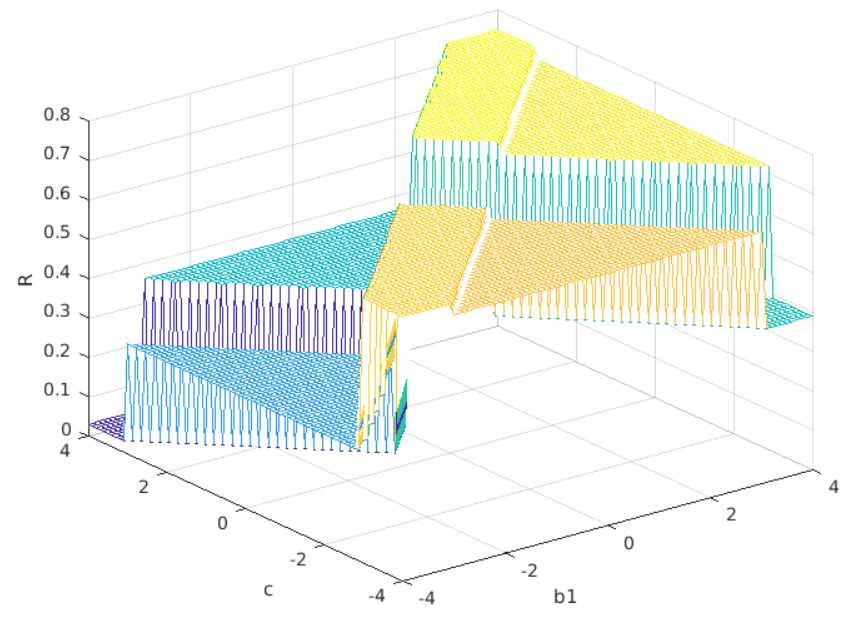} 
    \caption{Clean $0$-$1$ risk $R$ for case 1.}
    \label{fig: case1_Clean}
\end{figure}

Now, the set of classifiers obtained by minimizing the 0-1 risk is given in Figure \ref{fig: case1_Noisy}.
\begin{figure}[!h]
    \centering
\includegraphics[scale=0.4]{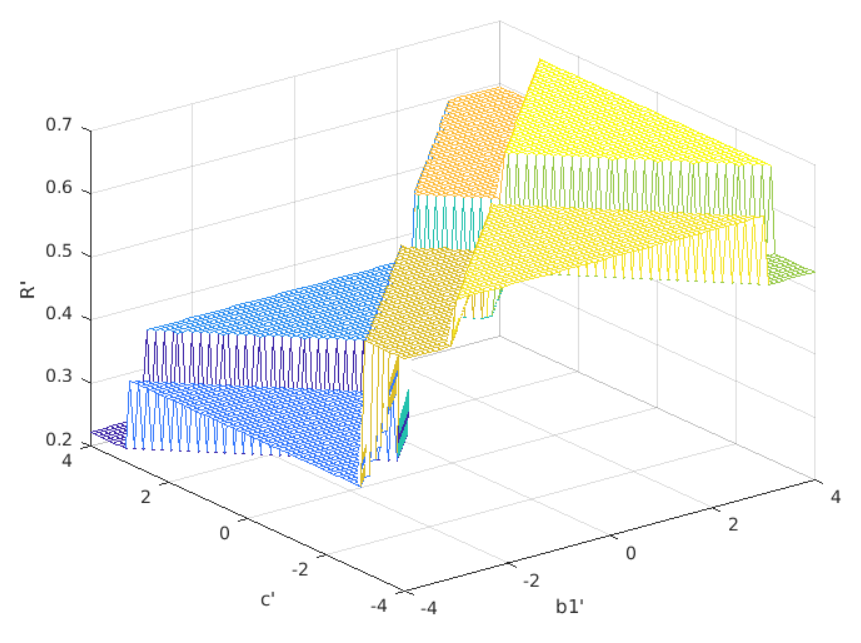}
    \caption{Noisy $0$-$1$ risk $R'$ for case 1.}
    \label{fig: case1_Noisy}
\end{figure}

 We can see that $f^{*}=-4x_{1}+x_{2}+4$ is a classifier which minimizes the clean as well as noisy risk. 
\begin{equation*}
    \text{\textbf{Case 2} }(\mathbf{x}^{(1)},y^{(1)}) = (\begin{bmatrix} +1 \\
    +1
    \end{bmatrix} , +1) \quad (\mathbf{x}^{(2)},y^{(2)}) = (\begin{bmatrix} +1 \\ -1
    \end{bmatrix}, -1)
    \end{equation*}
\begin{equation*}
    (\mathbf{x}^{(3)},y^{(3)}) = (\begin{bmatrix} -1 \\ -1
    \end{bmatrix}, -1) \quad (\mathbf{x}^{(4)},y^{(4)}) = (\begin{bmatrix} -1 \\ +1 
    \end{bmatrix}, +1)
\end{equation*} 

The set of classifiers obtained by minimizing the 0-1 risk is given in Fig \ref{fig: case2_Clean}.
\begin{figure}[!h]
    \centering
\includegraphics[scale=0.4]{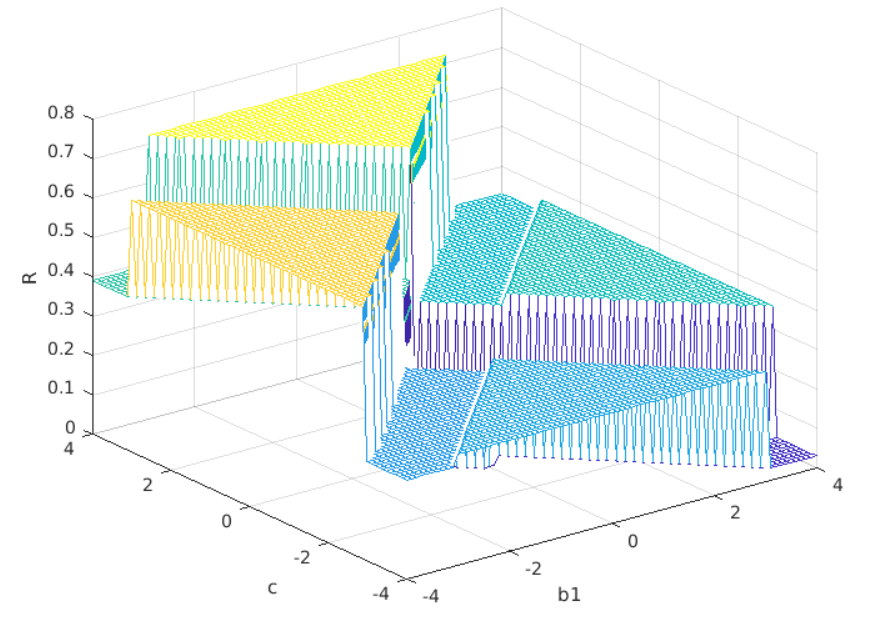} 
    \caption{Clean $0$-$1$ risk $R$ for case 2.}
    \label{fig: case2_Clean}
\end{figure}

Now, the set of classifiers obtained by minimizing the 0-1 risk is given in Figure \ref{fig: case2_Noisy}.
\begin{figure}[!h]
    \centering
\includegraphics[scale=0.4]{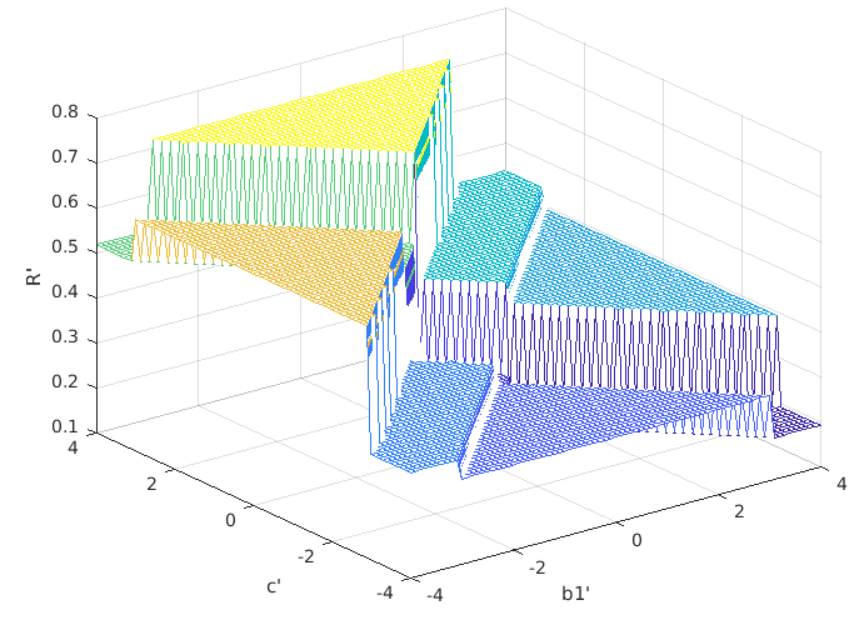}
    \caption{Noisy $0$-$1$ risk $R'$ for case 2.}
    \label{fig: case2_Noisy}
\end{figure}

 We can see that $f^{*}=4x_{1}+x_{2}-4$ is a classifier which minimizes the clean as well as noisy risk. 
 
\begin{equation*}
    \text{\textbf{Case 3}  }(\mathbf{x}^{(1)},y^{(1)}) = (\begin{bmatrix} +1 \\
    +1
    \end{bmatrix} , +1) \quad (\mathbf{x}^{(2)},y^{(2)}) = (\begin{bmatrix} +1 \\ -1
    \end{bmatrix}, -1) 
    \end{equation*}
\begin{equation*}
    (\mathbf{x}^{(3)},y^{(3)}) = (\begin{bmatrix} -1 \\ -1
    \end{bmatrix}, -1) \quad (\mathbf{x}^{(4)},y^{(4)}) = (\begin{bmatrix} -1 \\ +1 
    \end{bmatrix}, -1)
\end{equation*}  

The set of classifiers obtained by minimizing the 0-1 risk is given in Fig \ref{fig: case3_Clean}.
\begin{figure}[!h]
    \centering
\includegraphics[scale=0.4]{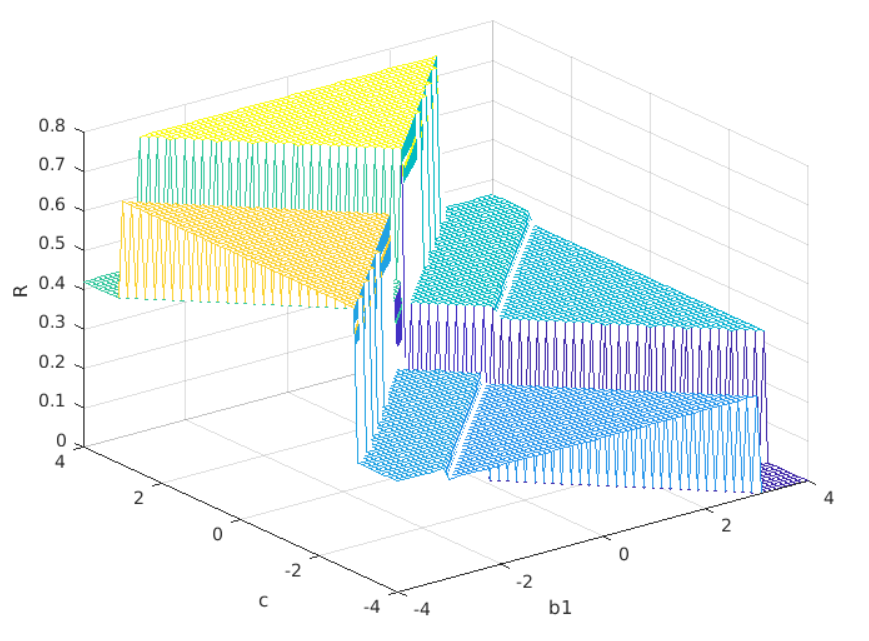} 
    \caption{Clean $0$-$1$ risk $R$ for case 3.}
    \label{fig: case3_Clean}
\end{figure}

Now, the set of classifiers obtained by minimizing the 0-1 risk is given in Figure \ref{fig: case3_Noisy}.
\begin{figure}[!h]
    \centering
\includegraphics[scale=0.4]{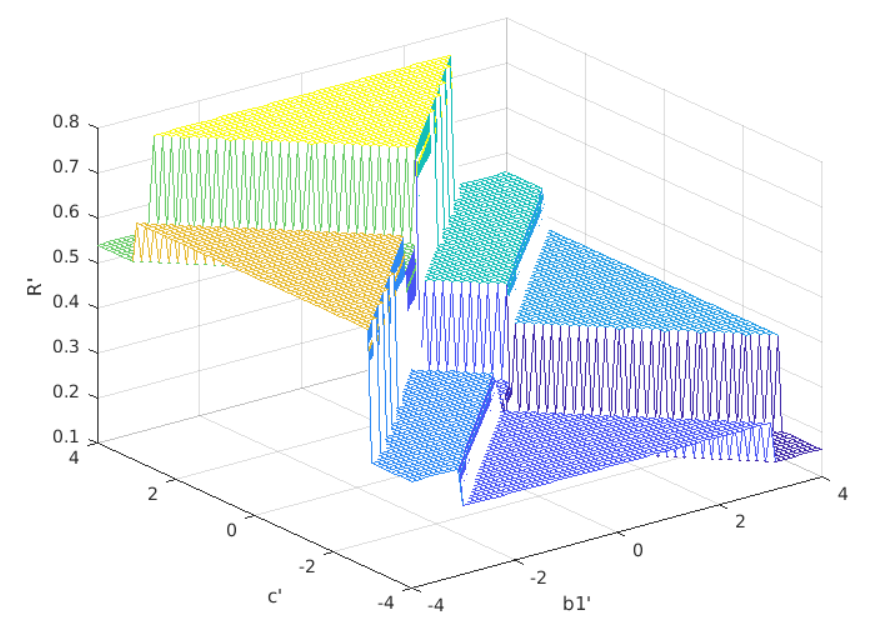}
    \caption{Noisy $0$-$1$ risk $R'$ for case 3.}
    \label{fig: case3_Noisy}
\end{figure}

We can see that $f^{*}=4x_{1}+x_{2}-4$ is a classifier which minimizes the clean as well as noisy risk. 

\begin{equation*}
    \text{\textbf{Case 4}  }(\mathbf{x}^{(1)},y^{(1)}) = (\begin{bmatrix} +1 \\
    +1
    \end{bmatrix} , -1) \quad (\mathbf{x}^{(2)},y^{(2)}) = (\begin{bmatrix} +1 \\ -1
    \end{bmatrix}, -1) 
    \end{equation*}
\begin{equation*}
    (\mathbf{x}^{(3)},y^{(3)}) = (\begin{bmatrix} -1 \\ -1
    \end{bmatrix}, -1) \quad (\mathbf{x}^{(4)},y^{(4)}) = (\begin{bmatrix} -1 \\ +1 
    \end{bmatrix}, -1)
\end{equation*}
The set of classifiers obtained by minimizing the 0-1 risk is given in Fig \ref{fig: case4_Clean}.

\begin{figure}[!h]
    \centering
\includegraphics[scale=0.4]{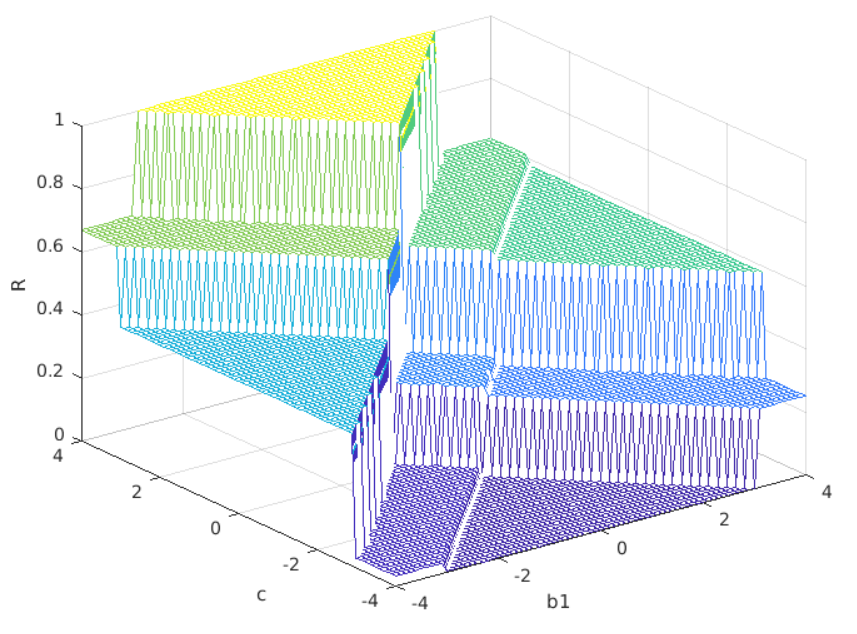} 
    \caption{Clean $0$-$1$ risk $R$ for case 4.}
    \label{fig: case4_Clean}
\end{figure}

Now, the set of classifiers obtained by minimizing the 0-1 risk is given in Figure \ref{fig: case4_Noisy}.
\begin{figure}[!h]
    \centering
\includegraphics[scale=0.4]{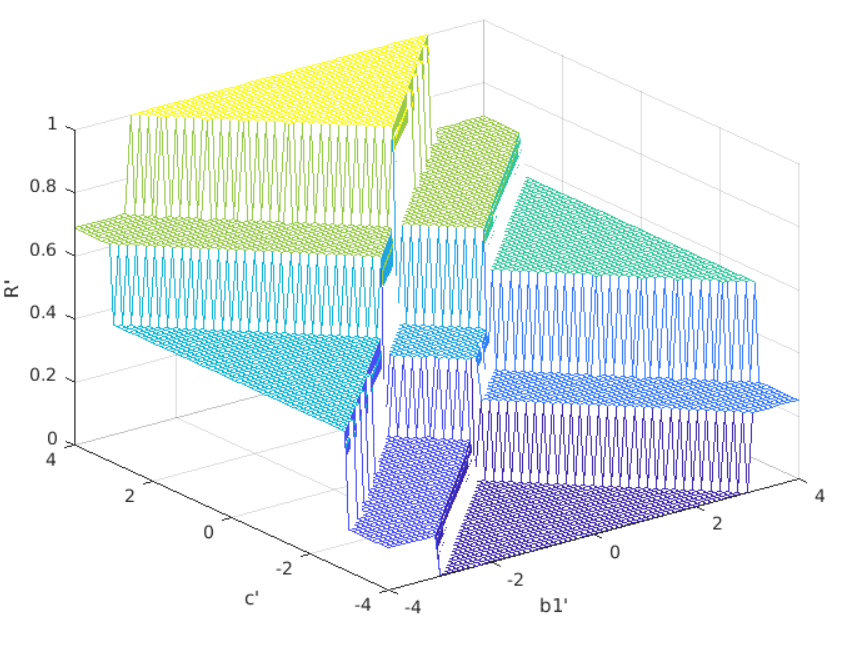}
    \caption{Noisy $0$-$1$ risk $R'$ for case 4.}
    \label{fig: case4_Noisy}
\end{figure}
 
 We can see that $f^{*}=x_{2}-3.9$ is a classifier which minimizes the clean as well as noisy risk.
 
 As seen in the above four cases, we conclude that the set of minimizers of clean $0$-$1$ risk and corrupted $0$-$1$ are same. And hence, the  classifiers learnt by minimizing the 0-1 risk noisy distributions are Asy-In attribute noise robust with $n=2$.

\end{proof}

\section{Details of counter examples}
\subsection{Counter example \ref{exa: 0-1_1d_sy-de}}
Consider a population of two data points of the form $(x,y)$ in 1-dimension as $(-1,1)$ and $(1,-1)$ with probability $0.25$ and $0.75$ and  consider the classifier of the form $f_{lin}(x) = bx+c$.
Then, the clean $0$-$1$ risk is given as follows:
$$R_{D}(f_{lin}) = 0.25\mathds{1}_{[-b+c<0]} + 0.75\mathds{1}_{[b+c>0]}$$
Minimizing $R_{D}(f_{lin})$ w.r.t. $b$ and $c$ gives us the optimal clean classifier and optimal clean risk as 
$$f_{lin}^* = (b^*,c^*) = (-1,-0.1),$$
$$ R_{D}(f_{lin}^*) = 0.$$
Next, we consider the Sy-De attribute noise corrupted risk with $p=0.4$ as follows:
\begin{eqnarray} \nonumber
R_{\tilde{D}}(\tilde{f}_{lin}) &=& (1-p)[0.25\mathds{1}_{[-\tilde{b}+\tilde{c}<0]} + 0.75\mathds{1}_{[\tilde{b}+\tilde{c}>0]}] \\
& & +p[0.25\mathds{1}_{[\tilde{b}+\tilde{c}<0]} + 0.75\mathds{1}_{[-\tilde{b}+\tilde{c}>0]}]
\end{eqnarray}
Minimizing $R_{\tilde{D}}(\tilde{f}_{lin})$ w.r.t. $ \tilde{b}$ and $\tilde{c}$ gives us the following optimal noisy  classifier and $0$-$1$ clean risk as
$$ \tilde{f}^*_{lin} = (\tilde{b}^*,\tilde{c}^*)=(1,-2)$$
$$ R_{D}(\tilde{f}_{lin}^*) = 0.25$$
Clearly, $R_{D}(f_{lin}^*) \neq R_{D}(\tilde{f}_{lin}^*)$ implying that $0$-$1$ loss need not be Sy-De attribute noise robust.

\subsection{Counter example \ref{exa: sq_2d_asy-in}}
Consider a population of 3 data points of the form $(x_1,x_2,y)$ in 2 dimensions as $(-1,1,1),(-1,-1,-1)$ and $(1,-1,+1)$ with probabilities as $(\frac{1}{4},\frac{1}{2},\frac{1}{4})$. We consider the classifier of the form $f_{lin}(\mathbf{x}) = b_1x_1+b_2x_2$. Then, the clean squared loss based risk is given as follows:
\begin{eqnarray*}
R_{D,l_{sq}}(f_{lin,sq}) &=& \frac{1}{4}(-b_1+b_2-1)^2 + \frac{1}{2}(-b_1-b_2+1)^2  \\
& &  + \frac{1}{4}(b_1-b_2-1)^2
\end{eqnarray*}
Minimizing $R_{D,l_{sq}}(f_{lin,l_{sq}})$ w.r.t. $b_1$ and $b_2$ leads to following system of equations:
\begin{eqnarray*}
2b_1-1 &=&0 \\
2b_2 - 1 &=& 0
\end{eqnarray*}
Solving the above system of equations gives us the optimal squared loss based clean classifier and clean $0$-$1$ risk as
$$f^*_{lin, l_{sq}} = (b_1^*,b_2^*) = (0.5,0.5)$$
$$R_{D}(f^*_{lin,l_{sq}}) = \frac{1}{2}$$
Next, we consider the Asy-In attribute noise corrupted risk in terms of $p_1$ and $p_2$ as follows:
\begin{footnotesize}
\begin{eqnarray*}
R_{\tilde{D},l_{sq}}(\tilde{f}_{lin,l_{sq}}) \\
= (1-p_1)(1-p_2)\bigg[ \frac{1}{4}(-\tilde{b}_1+\tilde{b}_2-1)^2 + \frac{1}{2}(-\tilde{b}_1-\tilde{b}_2+1)^2 \\
+\frac{1}{4}(\tilde{b}_1-\tilde{b}_2-1)^2 \bigg] + p_1p_2\Bigg[ \frac{1}{4}(\tilde{b}_1-\tilde{b}_2-1)^2 + \frac{1}{2}(\tilde{b}_1+\tilde{b}_2+1)^2 \\
+\frac{1}{4}(-\tilde{b}_1+\tilde{b}_2-1)^2 \bigg] + (1-p_1)p_2 \Bigg[ \frac{1}{4}(-\tilde{b}_1-\tilde{b}_2-1)^2 \\
+ \frac{1}{2}(-\tilde{b}_1+\tilde{b}_2+1)^2 +\frac{1}{4}(\tilde{b}_1+\tilde{b}_2-1)^2 \bigg] \\
+ p_1(1-p_2) \Bigg[\frac{1}{4}(\tilde{b}_1+\tilde{b}_2-1)^2 + \frac{1}{2}(\tilde{b}_1-\tilde{b}_2+1)^2 \\
+\frac{1}{4}(-\tilde{b}_1-\tilde{b}_2-1)^2 \Bigg]
\end{eqnarray*}
\end{footnotesize}
Minimizing $R_{\tilde{D},l_{sq}}(\tilde{f}_{lin,l_{sq}})$ w.r.t. $\tilde{b}_1, \tilde{b}_2$
leads to the following system of equations:
\begin{eqnarray*}
2\tilde{b}_1 &=& (1-2p_1) \\
2\tilde{b}_2 &=& (1-2p_2)
\end{eqnarray*}
Now, if we take the noise rate values as $p_1 = 0.1$ and $p_2 = 0.2$, then solving for above system of equations gives us the optimal squared loss based noisy linear classifier  and clean $0$-$1$ risk as
$$ \tilde{f}^*_{lin,l_{sq}} = (\tilde{b}_1^*,\tilde{b}_2^*) = (0.4,0.3)$$
$$R_{D}(\tilde{f}^*_{lin,l_{sq}}) = \frac{1}{4}$$
Clearly, $R_{D}(f_{lin,l_{sq}}^*) \neq R_{D}(\tilde{f}_{lin,l_{sq}}^*)$ implying that squared loss need not be Asy-In attribute noise robust.

\section{UCI dataset details}
In this section, we provide some details on the way we processed the datasets to obtain the experimental results. Number of features and the number of data points (number of negative and positive labelled separately) are provided in Table \ref{tab: dataset_details}. We provide individual pre-processing details for each dataset as follows:
\begin{itemize}
    \item \textbf{Vote:} This is a voting dataset with original size of 435 data points. Since missing entry in a cell meant that the person takes a neutral stand, we removed such instances to fit in the framework of binary valued attributes and finally used 232 data points. Label $+1$ corresponds to ``Democrat" and label $-1$ corresponds to ''Republican".
    \item \textbf{SPECT:} This dataset has information extracted from cardiac SPECT images with values as $0$ and $1$. To be consistent with the binary format, we replaced all $0$'s by $-1$ without loss of generality in both attributes and labels.
    \item \textbf{KR-vs-KP:} This dataset originally has 36 features but we removed feature number 15 as it had three categorical values and finally used the dataset with 35 attributes. Also, the attributes are processed as follows: ``f" replaced by ``+1", ``t" replaced by ``-1", ``n" replaced by ``+1", ``g" replaced by ``+1", ``l" replaced by ``-1". Finally,  label $+1$ corresponds to ``Won" and label $-1$ corresponds to ``Nowin".
\end{itemize}
\begin{table}[!h]
\centering
\begin{tabular}{|c|c|c|c|}
\hline
\textbf{S.no} & \textbf{Dataset name} & \textbf{n} & \textbf{m ($m_+$,$m_-$)} \\ \hline
\textbf{1} & Vote & 16 & 232 (124,108) \\ \hline
\textbf{2} & SPECT & 22 & 267 (212,55) \\ \hline
\textbf{3} & KR-vs-KP & 35 & 3196 (1569,1527) \\ \hline
\end{tabular}
\caption{Details about the number of features (n), number of total data points (m), number of positively labelled data points ($m_+$) and number of negatively labelled data points ($m_-$). }
\label{tab: dataset_details}
\end{table}
In Tables \ref{tab: real_accuracy} and \ref{tab: real_am}, we present the values which we used to generate the plots in Figure \ref{fig: real_data}. The results are averaged over 15 trials. We observe that at high noise rates, the theoretically proven robustness of squared loss doesn't work because of the finite samples.
\begin{table}[!h]
\centering
{\scriptsize
\begin{tabular}{|c|c|c|c|c|c|c|}
\hline
\textbf{Dataset} & \multicolumn{2}{c|}{\textbf{Vote}} & \multicolumn{2}{c|}{\textbf{SPECT}} & \multicolumn{2}{c|}{\textbf{KR-vs-KP}} \\ \hline
\textbf{(m,n)} & \multicolumn{2}{c|}{\textbf{(232 (124,108), 16)}} & \multicolumn{2}{c|}{\textbf{(267(212,55),22)}} & \multicolumn{2}{c|}{\textbf{(3196(1569,1527),35)}} \\ \hline
 & \textbf{Mean} & \textbf{SD} & \textbf{Mean} & \textbf{SD} & \textbf{Mean} & \textbf{SD} \\ \hline
\textbf{p} & 0.969 & 0.0218 & 0.732 & 0.0589 & 0.940 & 0.0109 \\ \hline
\textbf{0} & 0.966 & 0.0203 & 0.693 & 0.0746 & 0.938 & 0.0096 \\ \hline
\textbf{0.1} & 0.955 & 0.0363 & 0.680 & 0.0681 & 0.924 & 0.0106 \\ \hline
\textbf{0.2} & 0.865 & 0.0839 & 0.638 & 0.0797 & 0.900 & 0.0150 \\ \hline
\textbf{0.3} & 0.821 & 0.0941 & 0.625 & 0.0664 & 0.866 & 0.0193 \\ \hline
\textbf{0.35} & 0.694 & 0.0919 & 0.579 & 0.0955 & 0.795 & 0.0263 \\ \hline
\textbf{0.4} & 0.573 & 0.1041 & 0.505 & 0.0793 & 0.706 & 0.0393 \\ \hline
\end{tabular}}
\caption{Average (Mean) test accuracy along with standard deviation (SD) over 15 trials obtained by using squared loss based linear classifier learnt on Sy-De (noise rate p) attribute noise corrupted data. Even though squared loss is theoretically shown to be Sy-De noise robust,  it doesn't show good performance at high noise rates. This could be because the result in Theorem \ref{thm: nDi_sq}  is in expectation. In particular, finite sample size starts showing its effect at high noise rates and the performance deteriorates.}
\label{tab: real_accuracy}
\end{table}

\begin{table}[!h]
\centering
{\scriptsize
\begin{tabular}{|c|c|c|c|c|c|c|}
\hline
\textbf{Dataset} & \multicolumn{2}{c|}{\textbf{Vote}} & \multicolumn{2}{c|}{\textbf{SPECT}} & \multicolumn{2}{c|}{\textbf{KR-vs-KP}} \\ \hline
\textbf{(m,n)} & \multicolumn{2}{c|}{\textbf{(232 (124,108), 16)}} & \multicolumn{2}{c|}{\textbf{(267(212,55),22)}} & \multicolumn{2}{c|}{\textbf{(3196(1569,1527),35)}} \\ \hline
 & \textbf{Mean} & \textbf{SD} & \textbf{Mean} & \textbf{SD} & \textbf{Mean} & \textbf{SD} \\ \hline
\textbf{p} & 0.970 & 0.0207 & 0.665 & 0.0639 & 0.940 & 0.0111 \\ \hline
\textbf{0} & 0.967 & 0.0193 & 0.613 & 0.1214 & 0.937 & 0.0097 \\ \hline
\textbf{0.1} & 0.956 & 0.0373 & 0.605 & 0.1338 & 0.923 & 0.0109 \\ \hline
\textbf{0.2} & 0.868 & 0.0839 & 0.568 & 0.1353 & 0.899 & 0.0152 \\ \hline
\textbf{0.3} & 0.823 & 0.0950 & 0.586 & 0.1188 & 0.866 & 0.0186 \\ \hline
\textbf{0.35} & 0.698 & 0.0903 & 0.544 & 0.1542 & 0.794 & 0.0261 \\ \hline
\textbf{0.4} & 0.575 & 0.1045 & 0.540 & 0.1472 & 0.706 & 0.0399 \\ \hline
\end{tabular}}
\caption{Average (Mean) test  AM value along with standard deviation (SD) over 15 trials obtained by using squared loss based linear classifier learnt on Sy-De (noise rate p) attribute noise corrupted data.  Due to the imbalanced nature of SPECT dataset, AM is a more suitable evaluation metric.}
\label{tab: real_am}
\end{table}
\section{Additional examples}
\begin{example}
This is another example  which demonstrates that $0$-$1$ is robust to Asy-In attribute noise with $n=2$. Let the input data be two dimensional and the clean training set be \begin{equation*}
    (\mathbf{x}^{(1)},y^{(1)}) = (\begin{bmatrix} +1 \\
    +1
    \end{bmatrix} , +1) \quad (\mathbf{x}^{(2)},y^{(2)}) = (\begin{bmatrix} +1 \\ -1
    \end{bmatrix}, -1) \end{equation*} \begin{equation*}
    (\mathbf{x}^{(3)},y^{(3)}) = (\begin{bmatrix} -1 \\ -1
    \end{bmatrix}, -1)
\end{equation*}
and uniformly distributed with $\mathbf{x}^{(i)} = [x_{i1},x_{i2}], ~i=1,2,3$. Let the flipping probabilities of the first and second component be $p_{1}=0.12$ and $p_{2}=0.23$ respectively.  Let us consider the loss function to be the $0$-$1$ loss and the classifier be of the form $ f_{lin}(\mathbf{x}^{(i)}) = b_1x_{i1} +x_{i2} +c, i=1,2,3$. We calculate $f^{*}_{lin}$ as 
$$f_{lin}^{*} = \arg\min\limits_{f_{lin}} \mathbb{E}_{D}[l_{0-1}(y,f_{lin})]$$
We get a range of values for $b_1^*$ and $c^*$, we can choose one of the value which is $b_1^* = 1/3 = c^*$ that gives the risk to be 0. 

Now, for the corrupted case, we minimize the noisy $0$-$1$ classifier to obtain $\tilde{f^{*}_{lin}}(\tilde{x}^{(i)}) = \tilde{b}_1 \tilde{x}_{i1} +\tilde{x}_{i2} +\tilde{c},~ i=1,2,3$ given as below:
\begin{gather*}
    \tilde{f^{*}_{lin}} = \arg\min\limits_{\tilde{f^{*}}_{lin}} \mathbb{E}_{\tilde{D}}[l_{0-1}(y,\tilde{f}_{lin})] \\ 
    =\arg\min\limits_{\tilde{b}_1, \tilde{c}} \frac{1}{3}(1-p_{1})(1-p_{2})\Big[ \mathds{1}_{[\tilde{b}_1 + 1 + \tilde{c}< 0, y=1]} \\+ \mathds{1}_{[\tilde{b}_1 - 1 + \tilde{c}> 0, y=-1]}+\mathds{1}_{[-\tilde{b}_1 - 1 + \tilde{c}> 0, y=-1]}\Big] \\ 
    + \frac{1}{3}\Big[  p_{1})(p_{2})(\mathds{1}_{[-\tilde{b}_1 - 1 + \tilde{c}< 0, y=1]}+\mathds{1}_{[-\tilde{b}_1 + 1 + \tilde{c}> 0, y=-1]} \\ +\mathds{1}_{[\tilde{b}_1 + 1 + \tilde{c}> 0, y=-1]}\Big]  + \frac{1}{3}(1-p_{1})(p2)  \Big[ \mathds{1}_{[\tilde{b}_1 - 1 + \tilde{c}< 0, y=1]} \\ +\mathds{1}_{[\tilde{b}_1 + 1 + \tilde{c}> 0, y=-1]}+   \mathds{1}_{[-\tilde{b}_1 + 1 + \tilde{c}> 0, y=-1]}\Big] \\ + \frac{1}{3}(p_{1})(1-p_{2}) \Big[ \mathds{1}_{[-\tilde{b}_1 + 1 + \tilde{c}< 0, y=1]}+\mathds{1}_{[-\tilde{b}_1 - 1 + \tilde{c}> 0, y=-1]} \\+ \mathds{1}_{[\tilde{b}_1 - 1 + \tilde{c}> 0, y=-1]}\Big]
\end{gather*}
The minimizer for the above equation is calculated by plotting a graph (in MATLAB) given in Figure \ref{fig: exam_3}. Here, the values of $p_{1}$ and $p_{2}$ are taken to be 0.12 and 0.23 respectively. The same pattern is observed for all values of $p_1,p_2<0.5$
\begin{figure}
    \centering
\includegraphics[scale=0.5]{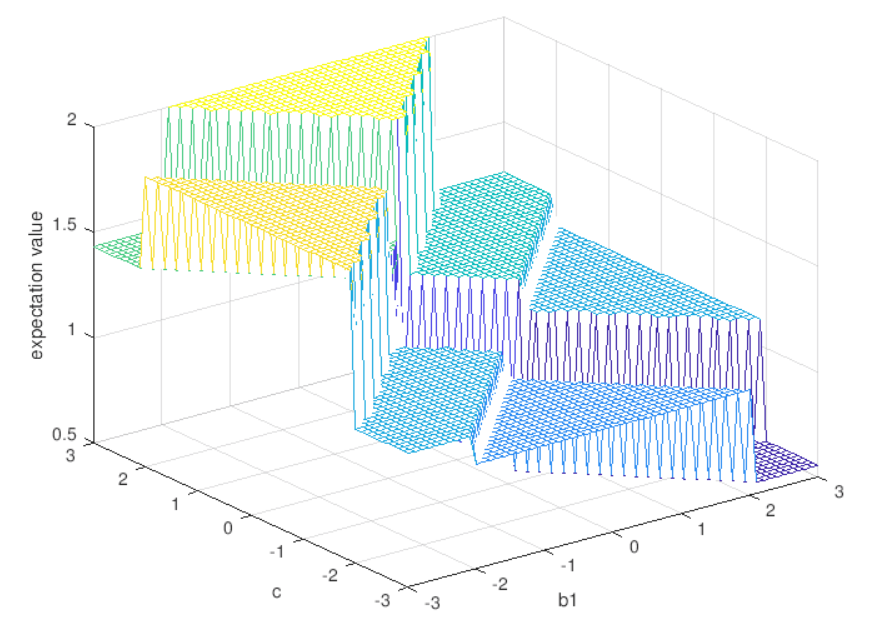} 
    \caption{Noisy $0$-$1$ risk is plotted on $z$ axis. The $x$ and $y$ axis labels are to be read as $\tilde{b}_1$ and $\tilde{c}$ as we are looking for the minimizers of noisy risk.}
    \label{fig: exam_3}
\end{figure}
The minimum value is obtained at $\tilde{b}_1^*=3$ and $\tilde{c}^*=-3$.

Comparing the clean $0$-$1$ risk  of the classifiers  $f^{*}_{lin}$ and $\tilde{f^{*}}_{lin}$, we observe that they are equal(0) and hence the $0$-$1$ loss function is attribute noise robust in this case. 
\end{example}

\end{document}